\documentclass[10pt,journal,compsoc]{IEEEtran}
\ifCLASSOPTIONcompsoc
\usepackage[nocompress]{cite}
\else
\usepackage{cite}
\fi
\ifCLASSINFOpdf
\usepackage[pdftex]{graphicx}
\else
\fi
\usepackage{amsmath}
\usepackage{url}

\usepackage{booktabs}
\usepackage{multirow}
\usepackage[caption=false,font=footnotesize]{subfig}
\usepackage{tikz}
\usetikzlibrary{positioning}
\usepackage{ragged2e}
\usepackage{setspace}

\usepackage{xspace}
\makeatletter
\DeclareRobustCommand\onedot{\futurelet\@let@token\@onedot}
\def\@onedot{\ifx\@let@token.\else.\null\fi\xspace}
\def\etal{\emph{et al}\onedot}

\begin{document}
	%
	\title{Anomaly Detection with Convolutional Autoencoders for Fingerprint\\ Presentation Attack Detection}
	%
	%
	%
	
	\author{Jascha Kolberg, Marcel Grimmer, Marta Gomez-Barrero, Christoph Busch
		\IEEEcompsocitemizethanks{\IEEEcompsocthanksitem Jascha Kolberg, Marcel Grimmer and Christoph Busch are with da/sec - Biometrics and Internet Security Research Group, Hochschule Darmstadt, Germany. 
			E-mails: \{firstname.lastname\}@h-da.de
			\IEEEcompsocthanksitem Marta Gomez-Barrero is with Hochschule Ansbach, Germany. \protect\\
			E-mail: \{firstname.lastname\}@hs-ansbach.de
			\IEEEcompsocthanksitem Marcel Grimmer, Christoph Busch are also with NTNU, Norway. \protect\\
			E-mail: \{firstname.lastname\}@ntnu.no}
	}

	\IEEEtitleabstractindextext{%
		\begin{justify}
			\begin{abstract}
				In recent years, the popularity of fingerprint-based biometric authentication systems significantly increased. However, together with many advantages, biometric systems are still vulnerable to presentation attacks (PAs). In particular, this applies for unsupervised applications, where new attacks unknown to the system operator may occur. Therefore, presentation attack detection (PAD) methods are used to determine whether samples stem from a bona fide subject or from a presentation attack instrument (PAI). In this context, most works are dedicated to solve PAD as a two-class classification problem, which includes training a model on both bona fide and PA samples. In spite of the good detection rates reported, these methods still face difficulties detecting PAIs from unknown materials. To address this issue, we propose a new PAD technique based on autoencoders (AEs) trained only on bona fide samples (i.e.\ one-class), which are captured in the short wave infrared domain.
On the experimental evaluation over a database of 19,711 bona fide and 4,339 PA images including 45 different PAI species, a detection equal error rate (D-EER) of 2.00\% was achieved. Additionally, our best performing AE model is compared to further one-class classifiers (support vector machine, Gaussian mixture model). The results show the effectiveness of the AE model as it significantly outperforms the previously proposed methods.   

			\end{abstract}
		\end{justify}
		
		}

	\maketitle

	\IEEEdisplaynontitleabstractindextext

	%
	\IEEEpeerreviewmaketitle

	\IEEEraisesectionheading{\section{Introduction}\label{sec:intro}}

\IEEEPARstart{N}{owadays}, 
we encounter biometric recognition systems in many places of our daily life. Applications range from high security border control to user convenient smartphone unlocking. Especially fingerprint recognition systems are long established and widely used~\cite{maltoni2009handbook}. 

However, biometric systems can be affected by external attacks as the capture device is exposed to the public. Those presentation attacks (PAs) are defined within ISO/IEC 30107-1~\cite{ISO-IEC-30107-1-PAD-Framework-160115} as a \lq\lq presentation to the biometric data capture subsystem with the goal of interfering with the operation of the biometric system\rq\rq. During execution, a presentation attack instrument (PAI), e.g.\ a fingerprint overlay, can be used to either impersonate someone else (i.e., impostor) or to avoid being recognised (i.e., identity concealer).
Summarising, the artefact that is used for a presentation attack is called PAI while different material combinations or recipies result in different PAI species.
As a consequence, biometric systems require automated presentation attack detection (PAD) modules in order to distinguish bona fide presentations from attack presentations~\cite{Marcel-HandbookPAD-ACVPR-2019}.

Since the periodic LivDet competitions started in 2009 for fingerprint~\cite{Yambay-ReviewFingerprintPAD-ACVPR-2019} and in 2013 for iris~\cite{Yambay-ReviewIrisPAD-ACVPR-2019}, PAD in general has attracted a lot of research. In parallel to those research efforts, more and more different materials are found or combined to create new species~\cite{Kanich-FingerprintPAIs-IWBF-2018}. On the one hand, older PAD methods might not detect new PAI species. On the other hand, it becomes much more challenging to collect diverse datasets in order to develop and evaluate (new) PAD approaches.
Being a binary classification problem (bona fide vs.\ PA), common PAD approaches are trained on both classes and hence perform only as good as the chosen training set. In this scenario, unknown attacks~\cite{Singh-SurveyUnknownFingerprintPAD-Arxiv-2020} present only in the test set can significantly trouble the classifier, as it requires good generalisation properties that are hard to achieve. In order to avoid re-training the classifier each time a new PAI species is created, one-class classifiers can be used~\cite{tax2002oneclass}. These models are solely trained on bona fide samples to detect anomalies in unseen data. They are especially designed to generalise much better than multi-class classifiers since all PAs are unknown to them. 

In this context, we propose to involve convolutional autoencoders for unknown fingerprint PAD. We test different architecture designs and show how the negative effect of outliers in the training set can be reduced in comparison to two-class classifiers. Finally, we benchmark the autoencoder against additional one-class classifiers to prove the soundness of our approach. The evaluation is carried out on data captured in the short wave infrared domain with over 24,000 samples, including 45 different PAI species. It should be noted, that the discussed design decisions should be generally applicable for other input data as well.

The remaining article is structured as follows: Section~\ref{sec:sota} summarises related work on fingerprint and one-class PAD. Our capture device is described in Section~\ref{sec:sensor} and Section~\ref{sec:pad} contains the autoencoder design and our proposed PAD method. In Section~\ref{sec:eval} we evaluate the experiments before Section~\ref{sec:conclusion} concludes our findings.

	
	
	\section{Related Work}
	\label{sec:sota}
	
This section reviews state-of-the-art approaches related to the contribution of this work. In the context of PAD, two different solutions exists: \textit{i) software-based}, where a deeper analysis of the existing data for authentication is carried out, and \textit{ii) hardware-based}, where new sensors are developed to capture additional data for PAD. Due to the high number of publications for fingerprint PAD within the last decade, we focus on hardware-based approaches in the first subsection and refer the reader to \cite{Sousedik-PAD-Survey-IET-BMT-2014, Marasco-PAD-SurveyFingerprint-CSUR-2014} for comprehensive surveys. On the other hand, most classifiers are trained on both classes, hence in the second subsection we present an overview of one-class PAD for other modalities as well. 
In order to evaluate the vulnerabilities of biometric systems to~PAs, the following metrics are defined within the ISO/IEC 30107-3 standard on biometric presentation attack detection - part 3: testing and reporting~\cite{ISO-IEC-30107-3-PAD-metrics-170227}:\\ 
\textbf{Attack Presentation Classification Error Rate (APCER)}: \lq\lq \emph{proportion of attack presentations using the same PAI species incorrectly classified as bona fide presentations}''.\\
\textbf{Bona fide Presentation Classification Error Rate (BPCER)}: \lq\lq \emph{proportion of bona fide presentations incorrectly classified as attack presentations}''.

\subsection{Hardware-based Fingerprint PAD}
\begin{table*}[t]
\centering
\caption{Hardware-based fingerprint PAD methods with their most relevant methodologies as performance and \linebreak the number of PAI species, PA samples, and bona fide samples.}
\begin{tabular}{cclcccc}
    \toprule
    \multirow{2}{*}{Year} & \multirow{2}{*}{Ref.} & \multirow{2}{*}{Description} & \multirow{2}{*}{Performance} & \#PAI & \#PA & \#BF \\
    & & & & species & samples & samples \\
    \midrule
    \multirow{2}{*}{2008} & \multirow{2}{*}{\cite{Rowe-Lumidigm-WP-2008}} & Multi-spectral wavelet transform & APCER = 0.9\% & \multirow{2}{*}{49} & \multirow{2}{*}{27,486} & \multirow{2}{*}{17,454} \\
    & & 430 nm, 530 nm, 630 nm + white light & BPCER = 0.5\% & \\
    \midrule
    \multirow{2}{*}{2011} & \multirow{2}{*}{\cite{Hengfoss-PAD-MultispectralFingerprint-FSI-2011}} & Multi-spectral & APCER = 0\% & \multirow{2}{*}{4} & \multirow{2}{*}{7-15} & \multirow{2}{*}{11-28}\\
    & & blanching effect, pulse & BPCER = 0\% & \\
    \midrule
    \multirow{2}{*}{2013} & \multirow{2}{*}{\cite{Drahansky-Fingerprint-PAD-BioMed-2013}} & Optical methods & APCER = 10\% & \multirow{2}{*}{N/A} & \multirow{2}{*}{N/A} & \multirow{2}{*}{N/A}\\
    & & pulse, pressure, skin reflections & BPCER $<$ 2\% & \\
    \midrule
    \multirow{2}{*}{2016} & \multirow{2}{*}{\cite{Darlow-OCTfingerPAD-AppliedOptics-2016}} & \multirow{2}{*}{OCT, double bright peaks + autocorrelation} & APCER = 0\% & \multirow{2}{*}{3} & \multirow{2}{*}{28} & \multirow{2}{*}{540}\\
    & &  & BPCER = 0\% & \\
    \midrule
    \multirow{10}{*}{2018} & \multirow{2}{*}{\cite{Gomez-Barrero-SWIR-SS-PAD-NISK-2018}} & \multirow{2}{*}{SWIR spectral signatures + SVM} & APCER = 5.7\% & \multirow{2}{*}{12} & \multirow{2}{*}{47} & \multirow{2}{*}{13}\\
    & &   & BPCER = 0\% & \\
    \cline{2-7}
     & \multirow{2}{*}{\cite{Tolosana-FingerPAD-CNN-SWIR-BIOSIG-2018}} & \multirow{2}{*}{SWIR + CNN} & APCER = 0\% & \multirow{2}{*}{12} & \multirow{2}{*}{47} & \multirow{2}{*}{13}\\
    & & & BPCER = 0\% & \\
    \cline{2-7}
    & \multirow{2}{*}{\cite{Keilbach-Fingerprint-LSCI-PAD-BIOSIG-2018}} & LSCI + SVM & APCER = 15.5\% & \multirow{2}{*}{32} & \multirow{2}{*}{225} & \multirow{2}{*}{545} \\
    & & BSIF, LBP, HOG, histogram & BPCER = 0.2\% & \\
    \cline{2-7}
    & \multirow{2}{*}{\cite{Hussein-LSCI-SWIR-CNN-FingerPAD-WIFS-2018}} & SWIR, LSCI & APCER = 0\% & \multirow{2}{*}{17} & \multirow{2}{*}{227} & \multirow{2}{*}{551}\\
    & & + patch-based CNN & BPCER = 0\% & \\
    \cline{2-7}
     & \multirow{2}{*}{\cite{Gomez-Barrero-FusionBATL-PAD-UBIO-2018}} & Weighted score fusion + SVM  & APCER = 6.6\% & \multirow{2}{*}{35} & \multirow{2}{*}{442} & \multirow{2}{*}{4,291}\\
    & & SWIR, LSCI, Finger vein & BPCER = 0.2\% & \\
    \midrule
    \multirow{10}{*}{2019} & \multirow{2}{*}{\cite{Liu-1DOCTPAD-ESA-2019}} & \multirow{2}{*}{OCT peak analysis} & APCER = 0\% & \multirow{2}{*}{4} & \multirow{2}{*}{24,000} & \multirow{2}{*}{12,000}\\
    & &  & BPCER = 0\% & \\
    \cline{2-7}
     & \multirow{2}{*}{\cite{Mirzaalian-LSCI-FingerPAD-2019}} &  LSCI CNN + LSTM & APCER $\le$ 0.14\% & \multirow{2}{*}{6} & \multirow{2}{*}{218} & \multirow{2}{*}{3,743} \\
    & & patch-based & BPCER $\le$ 0.11\% & \\
    \cline{2-7}
     & \multirow{2}{*}{\cite{Chugh-OCTFingerprintPAD-Arxiv-2019}} & OCT patch-CNN & APCER = 0.27\% & \multirow{2}{*}{8} & \multirow{2}{*}{357} & \multirow{2}{*}{3,413} \\
    & & (sensor captures no fingerprints) & BPCER = 0.2\% & \\
    \cline{2-7}
     & \multirow{2}{*}{\cite{Kolberg-LSCIBenchmarkFingerPAD-BIOSIG-2019}} &  LSCI benchmark + Fusion & APCER = 9.01\% & \multirow{2}{*}{35} & \multirow{2}{*}{442} & \multirow{2}{*}{4,291} \\
    & & BSIF, LBP, HOG, histogram & BPCER = 0.05\% & \\
    \cline{2-7}
    & \multirow{2}{*}{\cite{GomezBarrero-PAD-SWIR-LSCI-ICB-2019}} & Fusion of: SWIR + CNN and & APCER $\le$ 3\% & \multirow{2}{*}{35} & \multirow{2}{*}{442} & \multirow{2}{*}{4,291} \\
    & & LSCI + hand-crafted features & BPCER $\le$ 0.1\% & \\
    \midrule
    \multirow{8}{*}{2020} & \multirow{2}{*}{\cite{Kolberg-VeinBasedFingerPAD-Springer-2020}} &  Finger vein images & APCER = 11.61\% & \multirow{2}{*}{32} & \multirow{2}{*}{224} & \multirow{2}{*}{542} \\
    & & Gaussian pyramid + SVM & BPCER = 0.68\% & \\
    \cline{2-7}
     & \multirow{2}{*}{\cite{Tolosana-SWIR-PAD-CNNs-TIFS-2020}} & SWIR CNN fusion & APCER $\approx$ 7\% & \multirow{2}{*}{35} & \multirow{2}{*}{442} & \multirow{2}{*}{4,291} \\
    & & (pre-trained and from scratch) & BPCER = 0.1\% & \\
    \cline{2-7}
     & \multirow{2}{*}{\cite{GomezBarrero-MSSWIRCNN-CRC-2020}} &  SWIR CNN Fusion & APCER = 1.16\% & \multirow{2}{*}{41} & \multirow{2}{*}{3,310} & \multirow{2}{*}{8,214} \\
    & & 4D-input pre-processing & BPCER = 0.2\% & \\
    \cline{2-7}
     & \multirow{2}{*}{\cite{Kolberg-LSTM-FingerPAD-IJCB-2020}} & 1310 nm laser sequences & APCER = 3.71\% & \multirow{2}{*}{45} & \multirow{2}{*}{4,339} & \multirow{2}{*}{17,730} \\
     & & LSTM + CNN benchmark & BPCER = 0.2\% & & &  \\
    \bottomrule
\end{tabular}
\label{tab:sotaHW}
\end{table*}

Similar to other pattern recognition tasks, PAD benefits from information captured by additional sensors. This information is then analysed with dedicated software. To that end, an overview of hardware-based state-of-the-art fingerprint PAD methods is presented in Table~\ref{tab:sotaHW}. 

One of the most reliable methods for fingerprint PAD is based on optical coherence tomography (OCT)\cite{oct} sensors, which capture a 3D model of the fingertip up to two millimeter underneath the skin. In addition to PAD, this scan can be used to recover worn-out fingerprints, since it includes the inner fingerprint as well. Hence, it also reveals overlaying PAIs as well as full fake fingers. Using OCT scanners, Darlow \etal~\cite{Darlow-OCTfingerPAD-AppliedOptics-2016} detected double bright peaks in gelatin overlays and analysed the autocorrelation for gelatin full fingers. Their setup achieves a 100\% detection accuracy on a database with 568 samples. Also Liu \etal~\cite{Liu-1DOCTPAD-ESA-2019} analyse the peaks of OCT scans. They discover that 1D depth scans of bona fide samples contain exactly two peaks while one appears prior the maximum peak. Thus, they apply a threshold to successfully distinguish between bona fides and PAs. Training a convolutional neural network (CNN) on overlapping patches of a depth B-scan, Chugh \etal~\cite{Chugh-OCTFingerprintPAD-Arxiv-2019} report a detection accuracy close to 100\%. However, the utilised capture device does not acquire the fingerprint for biometric recognition purpose. An extensive review on OCT for fingerprint PAD is published by Moolla \etal~\cite{Moolla-OCTFingerprintPAD-Survey-HoAS-2019}. It should be noted that the high costs of OCT scanners are an explicit disadvantage in contrast to other methods.

Another approach utilises different illumination sources to collect additional PAD data. Rowe \etal~\cite{Rowe-Lumidigm-WP-2008} developed the first multi-spectral fingerprint capture device in 2008. Their sensor captures the fingerprint in white, blue, green, and red illumination with a twofold goal: \textit{i)}~improving the recognition process, and \textit{ii)} detection of PAIs. The authors prove the suitability of their design on a massive dataset of nearly 45,000 samples comprising 60\% PAs. In a similar approach, Hengfoss \etal~\cite{Hengfoss-PAD-MultispectralFingerprint-FSI-2011} analysed the reflections for all wavelengths between 400 nm and 1650 nm on the blanching effect (i.e., the finger is pressed against a surface such that the blood is squeezed out). They observe that these dynamic effects only occur for bona fide presentations and neither for PAIs nor for cadaver fingers. Additionally, they measured the pulse of the finger but conclude that it takes much longer and is less suited for PAD. Further optical methods for pulse, pressure, and skin reflections are presented by Drahansky \etal~\cite{Drahansky-Fingerprint-PAD-BioMed-2013}. Their experiments show that skin reflections in the evaluated wavelengths of 470 nm, 550 nm, and 700 nm outperform the other two methods. In another approach, Kolberg \etal~\cite{Kolberg-VeinBasedFingerPAD-Springer-2020} visualise vein patterns by placing 940 nm LEDs above the finger. Using Gaussian pyramids, they are able to detect fingeprint PAIs since they usually do not include a vein pattern. However, for thin and transparent overlay attacks the bona fide veins still remain visible, which limits detection capabilites for overlay PAIs.

More recent publications focus on the short wave infrared (SWIR) spectrum between 900 nm and 1700 nm, which is not visible for the human eye but can be captured by adequate cameras. Gomez-Barrero \etal~\cite{Gomez-Barrero-SWIR-SS-PAD-NISK-2018} utilise the spectral signature between different wavelengths for fingerprint PAD. Working with a rather small database, they show that most materials reflect the illumination in a different way than human skin. A subsequent study~\cite{Tolosana-FingerPAD-CNN-SWIR-BIOSIG-2018} further improves PAD performance on those 60 samples with the use of a CNN. Moreover, by fine-tuning two pre-trained CNNs and training a small residual network from scratch, Tolosana \etal~\cite{Tolosana-SWIR-PAD-CNNs-TIFS-2020} showed that deep learning approaches perform much better than spectral signatures for bigger datasets. Additionally, the results reveal that the small residual network trained from scratch outperforms the fine-tuned VGG19 and MobileNet CNNs, for user-convenient scenarios requiring a low BPCER. Another extensive benchmark~\cite{GomezBarrero-MSSWIRCNN-CRC-2020} tests two additional CNNs and adds an advanced pre-processing layer to them. This layer is trained on the given dataset to pre-process a 4-channel SWIR image for usage in 3-channel CNNs, which significantly improves PAD performance in contrast to the manual pre-processing used in \cite{Tolosana-SWIR-PAD-CNNs-TIFS-2020}. 

On the other hand, the technique of laser speckle contrast imaging (LSCI)~\cite{Senarathna-LSCI-IEEERevBiomedEng-2013} is able to visualise blood movement underneath the skin. For this purpose, a laser illuminates the desired area and a sequence (i.e., 1 second) of images is captured. Since this laser slightly penetrates the skin, subtle movements within blood tissues change the reflected speckle pattern over time~\cite{Vaz-LSCI-IEEERevBiomedEng-2016}. Utilising this principle for fingerprint PAD, Keilbach \etal~\cite{Keilbach-Fingerprint-LSCI-PAD-BIOSIG-2018} compute the temporal contrast in order to obtain a single LSCI image for feature extraction. Those handcrafted features (e.g, LBP, BSIF) are then classified by support vector machines (SVMs). This approach was later benchmarked in \cite{Kolberg-LSCIBenchmarkFingerPAD-BIOSIG-2019} with eight additional classifiers on a larger dataset in order to evaluate the best PAD performance by fusing different schemes. However, similar to the work on vein patterns, thin and transparent overlays are often wrongly classified as bona fide. In the case that the material of the PAI is thin enough for the laser to still penetrate into the skin below, bona fide properties are captured and thus the PAI is not detected. Finally, Mirzaalian \etal~\cite{Mirzaalian-LSCI-FingerPAD-2019} applied deep learning methods on these laser sequences. Next to more traditional CNNs, the authors propose the usage of long short-term memory (LSTM) networks, which are able to remember a temporal state and can directly process the temporal information within sequences. The results show a slight advantage of the LSTM towards the four CNNs tested. A more extensive benchmark on LSTMs and CNNs in \cite{Kolberg-LSTM-FingerPAD-IJCB-2020} comes to the conclusion that both temporal analysis of the LSTMs and spatial analysis of some CNNs are partly complementary and detect different PA samples.

Given the promising concepts of SWIR and LSCI data for fingerprint PAD, fusions of both approaches have been published in \cite{Gomez-Barrero-FusionBATL-PAD-UBIO-2018, Hussein-LSCI-SWIR-CNN-FingerPAD-WIFS-2018, GomezBarrero-PAD-SWIR-LSCI-ICB-2019}. These multimodal approaches prove that PAD benefits from additional sensors. The weaknesses of one technology can be covered by another and the combination of different methods significantly improves the overall detection accuracy. Additionally, fused systems are more robust against unseen PAI species in the test set.

\subsection{One-class Presentation Attack Detection}

\begin{table}[t]
    \centering
    \caption{One-class PAD methods across different modalities.}
    \begin{tabular}{cccl}
    \toprule
    Year & Ref.\ & Modality & Description \\
    \midrule
    \multirow{2}{*}{2016} & \multirow{2}{*}{\cite{Ding-OneClassFingerPAD-WIFS-2016}} & \multirow{2}{*}{Fingerprint} & multiple OC-SVMs\\
    & & & + PA refining \\
    \midrule
    \multirow{2}{*}{2018} & \multirow{2}{*}{\cite{nikisins2018effectiveness}} & \multirow{2}{*}{Face} & OC-SVM + OC-GMM \\
    & & & vs.\ two-class versions \\
    \midrule
    \multirow{2}{*}{2019} & \multirow{2}{*}{\cite{Nikisins-AutoencoderFacePAD-ICB-2019}} & \multirow{2}{*}{Face} & Autoencoder + \\
    & & & multi-layer perceptron \\
    \midrule
    \multirow{2}{*}{2019} & \multirow{2}{*}{\cite{engelsma2019}} & \multirow{2}{*}{Fingerprint} & Three GANs based on \\
    & & & DCGAN architecture~\cite{radford2015unsupervised} \\
    \bottomrule
    \end{tabular}
    \label{tab:oc-sota}
\end{table}

Unlike traditional classification problems, the motivation behind one-class classifiers is learning the structure of data samples belonging to a single class. Therefore, in case of PAD, one-class classifiers are trained only on bona fide samples. New and unseen samples are classified as PAs if their structure differs from those bona fide samples used in the training phase. In this context, the main challenge is to find an optimal threshold to ensure that sophisticated PAs can still be distinguished from those bona fides that deviate from normality. Due to the environmental conditions and interaction factors (data subject with respect to the capture device) a significant intra-class variation for the bona fide class must be expected. Since the majority of published PAD approaches are based on two-class classification, this section reviews one-class publications across modalities as summarised in Table~\ref{tab:oc-sota}. Due to the different modalities and datasets used, a comparison of performance metrics is not included.

Generally, one-class classifiers can be split into generative and non-generative approaches~\cite{nikisins2018effectiveness}. Generative methods aim to approximate the distribution function of the bona fides (e.g. a Gaussian model). Non-generative approaches focus on learning an optimal hypersphere that defines a decision boundary to separate bona fides from PAs. 

One non-generative fingerprint PAD approach has been presented by Ding and Ross~\cite{Ding-OneClassFingerPAD-WIFS-2016}, who introduced an ensemble of multiple one-class support vector machine (OC-SVM) classifiers, each of which is trained on different feature sets. The main goal of all OC-SVMs is to find the smallest possible hypersphere around the majority of training samples. Once the boundaries of the hyperspheres are found, they are refined using a small number of PA samples. Finally, in order to obtain a single prediction, the scores of all OC-SVMs are fused by majority voting. With regard to unknown attacks not seen in the training phase, the authors reported an averaged APCER of 15.3\% vs. an averaged BPCER of 10.8\% on the LivDet 2011 database~\cite{LivDet2011-PAD-Fingerprint-ICB-2012}.

Another non-generative approach for face PAD has been proposed by Nikisins \etal~\cite{Nikisins-AutoencoderFacePAD-ICB-2019}, who use a combination of pre-trained autoencoders (AEs) and a simple multi-layer perceptron (MLP) for the final classification. The AEs are used to extract features from multi-channel input data, which in this case is a stack of greyscale, near-infrared, and depth facial images (BW-NIR-D). Each of the AEs are only trained on bona fide samples, thereby learning the appearance of real faces. Instead of collecting a lot of training data, Nikisins \etal use transfer learning techniques to transmit the knowledge of facial images from the RGB to the BW-NIR-D domain. The CNN model was pretrained on the CelebA~\cite{CelebaFaceDB-2018} database containing RGB facial images, which they fine-tuned on the Wide Multi-Channel Presentation Attack database (WMCA)~\cite{WMCA-PAD-DB-2019}. Only the subsequent MLP is trained on both bona fide and PA samples for the final classification of the face images. The authors report a BPCER of 7.3\% vs. an APCER of 1\%. 

In another work on face PAD, Nikisins \etal~\cite{nikisins2018effectiveness} implemented and tested both one-class Gaussian mixture models (OC-GMM) (generative) and OC-SVMs (non-generative), benchmarking their results, with two-class approaches as well. For their experiments, the authors employed an aggregated database as a composition of three publicly available databases: Replay-Attack \cite{chingovska2012effectiveness}, Replay-Mobile~\cite{costa2016replay}, and MSU MFSD~\cite{wen2015face}. Their results show a significant better detection performance for the OC-GMM approach compared to the OC-SVM. Particularly, they emphasise the ability of the OC-GMM to have better generalisation properties to unknown attack types as compared to the two-class classifiers and the OC-SVMs. Both models were trained on the image quality metric features introduced in \cite{wen2015face} and \cite{Galbally-IQApad-TIP-2014}.

\begin{figure}[t]
    \centering
    \includegraphics[width=\linewidth]{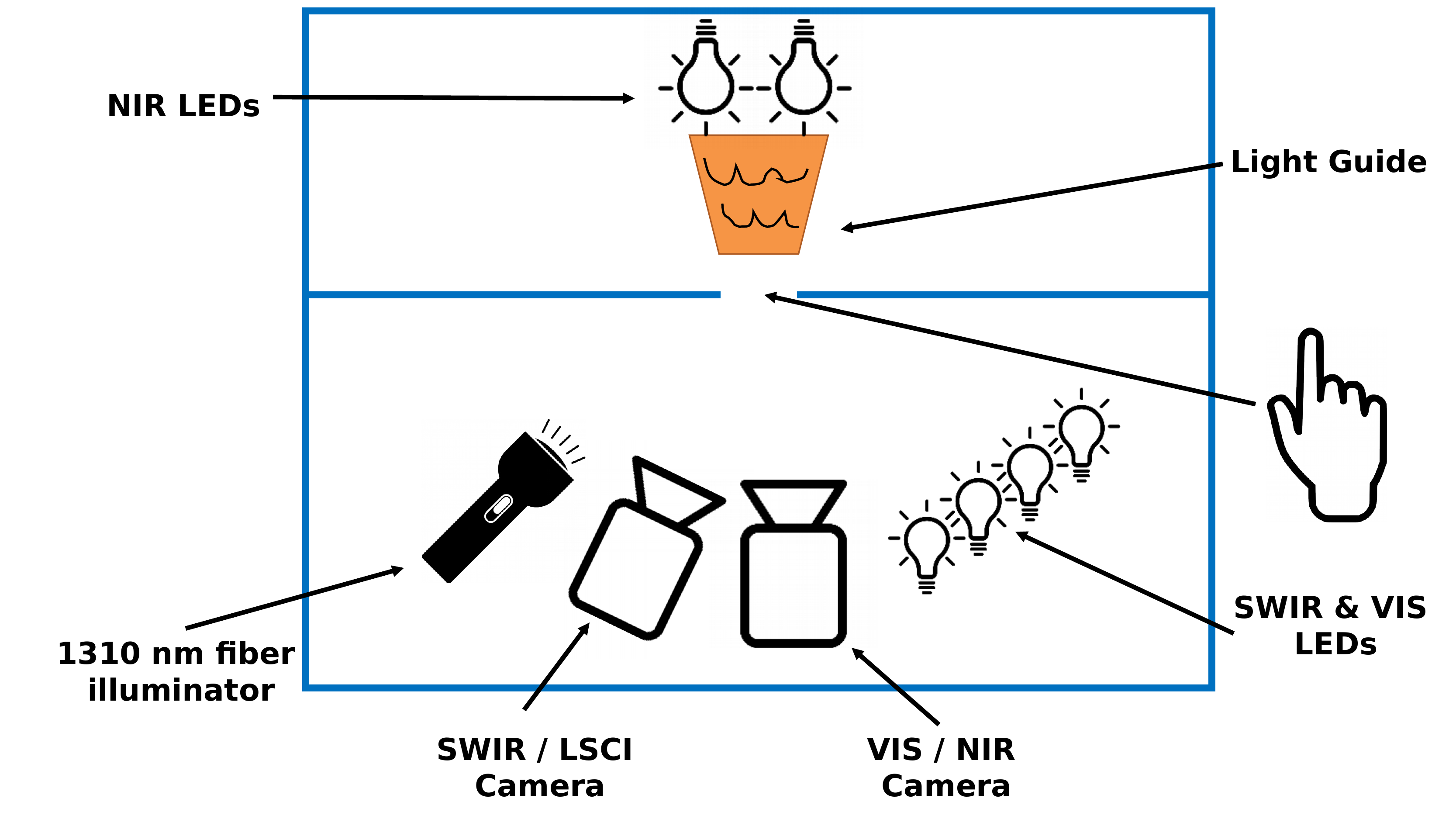}
    \caption{The capture device is a closed box with only one free slot for the finger. Two cameras and multiple illuminations are able to capture the fingerprint and additional PAD data.}
    \label{fig:sensor}
\end{figure}
Lastly, Engelsma and Jain~\cite{engelsma2019} present another one-class approach using generative adversarial networks (GANs) for fingerprint PAD. Specifically, they trained three different GAN models using the DCGAN architecture proposed by Radford \etal~\cite{radford2015unsupervised}. As part of their work, they collected a dataset comprising 12 different PAIs and 11,800 bona fide samples. The experimental evaluation reports an APCER of 15.6\% for a BPCER of 0.2\%.

	\section{Capture Device}
	\label{sec:sensor}
	
The camera-based fingerprint capture device~\cite{Spinoulas-BATLDatasetFingerPAD-Arxiv-2020} that was used for data collection is depicted in Fig.~\ref{fig:sensor}. One camera (Basler acA1300-60gm) takes finger photos in the visible spectrum to extract the fingerprint for legacy compatibility. This camera is also able to capture finger vein images, when only the near-infrared (NIR) LEDs above the finger are switched on. A second camera (100 fps Xenics Bobcat 320) captures PAD data in wavelengths between 900 nm and 1700 nm. Both cameras are placed in a closed box next to multiple illumination sources with only one finger slot at the top. Once a finger is placed on this slot, all ambient light is blocked and only the desired wavelengths illuminate the finger. The invisible SWIR wavelengths of 1200 nm, 1300 nm, 1450 nm, and 1550 nm are especially suited for PAD because all skin types in the Fritzpatrick scale~\cite{Fritzpatrick-SkinTypes-Dermatology-1988} reflect in the same way as shown by Steiner \etal~\cite{Steiner-facePADswir-ICB-2016} for face PAD. Hence, SWIR images are captured in each of these wavelengths.
Additionally, a 1310 nm laser diode illuminates the finger area and a sequence of 100 frames is collected within one second. Stemming from biomedical applications, this laser sequence is used to image and monitor microvascular blood flow~\cite{Senarathna-LSCI-IEEERevBiomedEng-2013}. Since the laser scatters differently when penetrating human skin in contrast to artificial PAIs, this technique qualifies for PAD as well.

\begin{figure}[t]
    \centering
    \input{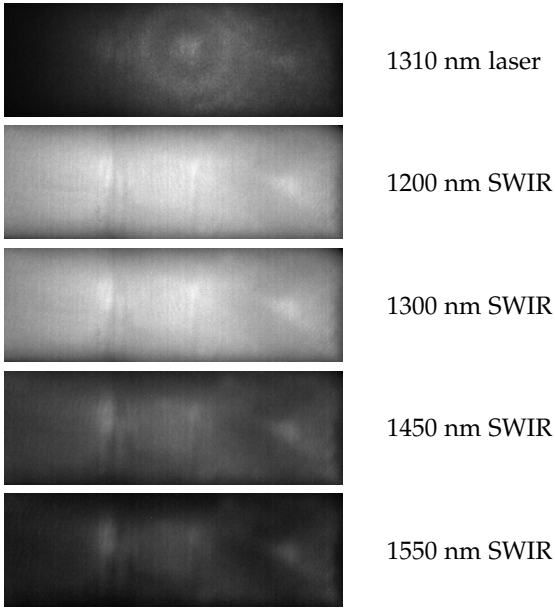}
    \caption{Bona fide samples acquired at five different wavelengths.}
    \label{fig:samples}
\end{figure}
Example frames of a bona fide presentation acquired at the aforementioned wavelengths are shown in Fig.~\ref{fig:samples}. For the laser sequence data, only one frame is depicted since the subtle temporal changes are not visible in steady pictures. Nevertheless, we can recognise a circle where the laser focuses the finger. On the other hand, the LEDs achieve a much more consistent illumination for the SWIR images, where the skin reflections get darker for increasing wavelengths.
The region of interest for all samples comprises $100\times 300$ pixels due to the fixed size of the finger slot.

	\section{Proposed PAD Method}
	\label{sec:pad}
	This Section introduces our one-class fingerprint PAD scheme based on a convolutional autoencoder, which is described in Section~\ref{sec:CAE}. Since AEs measure the reconstruction error, this concept is subsequently discussed in detail in Section~\ref{sec:reconstruction_error}. Finally, this scheme is combined with fingerprint PAD in Section~\ref{sec:pad_scheme}.

\subsection{Convolutional Autoencoder}
\label{sec:CAE}
A convolutional autoencoder is a neural network optimised to copy its input data. The model consists of two components: the encoder function $h=f(x)$ and the decoder function $x'=g(h)$, both of which are implemented as a multi-layer CNN. This means that the AE maps an input image $x$ to an output image $x'$. The output $h$ of the encoder function $f$ is a lower dimensional latent representation of the original image $x$. Out of this latent variable, the decoder function $g$ tries to reconstruct the original image $x$. In order to force the model to learn correct parameters for decoding the latent representation, a loss function needs to be minimised:
\begin{equation}
    L \left ( x,g\left ( f\left (x\right ) \right ) \right )
    \label{eq:loss_function}
\end{equation}
This loss function penalises $g(f(x))$ if it is dissimilar to $x$. The choice of the loss function thus plays a decisive role in the performance of convolutional AEs. In order to increase the efficiency of the learning process, the loss value can be calculated on a randomly selected subset called \emph{Batch}. However, one important requirement is to design the architecture of an AE in an undercomplete way. In other words, the dimension of $h$ needs to be smaller than the original dimension of input $x$. This forces the AE to only extract the most relevant features from the training data. Furthermore, it prevents the model to be in danger of learning the identity function $id(x)=x$~\cite{autoencoder}. 
Once the model is trained, it is able to encode and reconstruct images $x'$, which resemble the training data. In case of an input image that is dissimilar to the ones involved in training, the reconstruction fails and leads to a high reconstruction error (see Eq.~\eqref{eq:loss_function}). The high input sensitivity of an AE can be exploited to detect images that differ from the ones being used during training. For this reason, AEs became very popular in the field of anomaly detection (e.g. \cite{Nikisins-AutoencoderFacePAD-ICB-2019}, \cite{ishii2019}). Transferred to the domain of fingerprint PAD, the AE is only trained on bona fide samples. Later, the model can be used to detect unknown PAs by comparing the reconstruction error against a threshold.

\subsection{Reconstruction Error (RE)}
\label{sec:reconstruction_error}
A common approach to compute the reconstruction error is to use the mean squared error (MSE)~\cite{berger2013} as loss function, which is defined as
\begin{equation}
\label{eq:mse}
\begin{split}
    L(x, x') & = \frac{1}{B}\sum_{j=1}^B \frac{1}{WHI} \sum_{w=1}^W \sum_{h=1}^H \sum_{i=1}^I (x^j_{whi}-x'^j_{whi})^2 \\ & = \frac{1}{B}\sum_{j=1}^B \frac{1}{WHI} \sum_{w=1}^W \sum_{h=1}^H \sum_{i=1}^I e^j_{whi}(x, x')
\end{split}
\end{equation}
where $B, W, H$ and  $I$ denote the number of data samples involved in one batch iteration, the image width, height and the number of input channels of an input image $x$. The usage of MSE is convenient since it is easy understandable and often pre-implemented. However, there is also a major drawback in case of random noise occurring in the data. Since the calculation of the MSE involves squaring the difference between every pixel of the input image, single outliers have a huge impact on the reconstruction error. This inevitably leads to an increased rate of bona fide samples erroneously classified as PAs. This lack of robustness against outliers is a well known challenge in the deep learning domain and is referred to as \emph{robust estimation}~\cite{rousseeuw2011}. The idea of increasing the robustness of an AE model for anomaly detection was studied by Ishii and Takanashi~\cite{ishii2019}, who introduced a weighted version of the MSE (wMSE):
\begin{equation}
   L_{Ishii}(x, x') = \frac{1}{B}\sum_{j=1}^B w^j \cdot mse^j(x, x')
\end{equation}
where
\begin{equation}
    mse^j(x, x') = \frac{1}{WHI} \cdot \sum_{w=1}^W \sum_{h=1}^H \sum_{i=1}^I e^{j}_{whi}(x,x')
\end{equation}
and $w^j$ is defined as
\begin{equation}
         w^j=\left\{\begin{array}{ll} 1, & mse^j(x, x') \leq C \\
         0, & mse^j(x, x') > C \end{array}\right. .
\end{equation}
Here, $C$ refers to the $\alpha$-th quantile of $mse=[mse^1, \dots ,mse^B]$. The approach of Ishii and Takanashi ignores training samples during the optimisation process as soon as their measured MSE exceeds a defined threshold $C$. Translated to the problem of fingerprint PAD, that means that a certain percentage of bona fides is ignored during the training phase. The authors state that their proposed loss function is useful to cope with unknown outliers within the training set, since they will not distort the resulting model. Unknown outliers can occur, for example, if the data is not labelled. Therefore, it is difficult to differentiate them from normal data samples. However, in our case the training data contains no PAs. That means that excluding bona fide samples from the training process could potentially lead to a loss of information.

For that reason, the proposed loss function of Ishii and Takanashi is adjusted within this work. The main idea is to integrate the weight factor such that it excludes pixel values that the AE is systematically not able to reconstruct. In other words, this means that the AE is optimised to reconstruct the most meaningful areas of the images while ignoring random noise. The adjusted loss function is defined as follows:

\begin{equation}
\label{eq:proposed_wmse}
    L_{Prop}(x, x') = \frac{1}{B} \cdot \sum_{j=1}^B \frac{1}{WHI} \sum_{w=1}^W \sum_{h=1}^H \sum_{i=1}^I w^j_{whi} e^{j}_{whi}(x, x')
\end{equation}
with 
\begin{equation}
\label{eq:weight_definition_wMSE}
         w^j_{whi}=\left\{\begin{array}{ll} 1, & e^{j}_{whi}(x, x') \leq mse^j(x, x') + C \cdot std^j \\
         0, & e^{j}_{whi}(x, x') > mse^j(x, x') + C \cdot std^{j} \end{array}\right. 
\end{equation}
and 

\begin{equation}
    std^{j}  = \sqrt{\frac{1}{WHI}\cdot \sum_{w = 1}^W \sum_{h=1}^H \sum_{i=1}^I \left(e^{j}_{whi}(x, x') - mse^j(x, x')\right)^2}
\end{equation}

Generally speaking, every pixel value is compared to a threshold that is a linear combination of both mean and standard deviation of the squared error. Thus, exceeding pixels are ignored and contrary to the MSE, it is assumed that this approach prevents random noise from increasing the overall reconstruction error of the bona fide samples. The remaining challenge however consists in finding the optimal constant value of $C$. By choosing a too low threshold, the model might tend to over-generalise such that decisive patterns that are important for distinguishing between bona fides and PAs are not extracted anymore. On the other hand, if $C$ is too high, noisy data might be involved in both training and testing, which leads to a less robust model and consequently increases error rates. This problem is related to the typical trade-off between bias and variance.

\begin{figure*}[t]
    \centering
    \includegraphics[width=\textwidth]{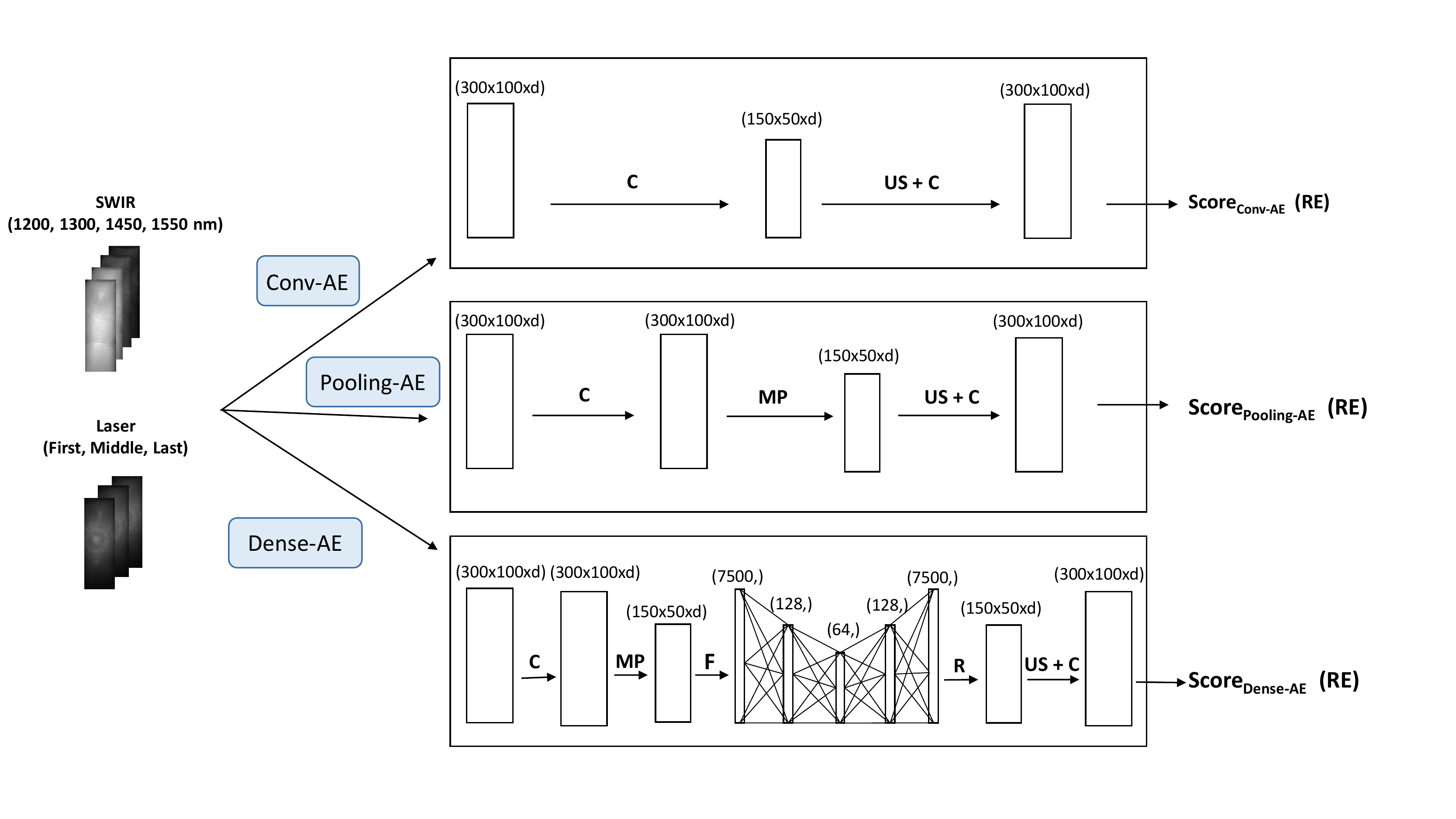}
    \caption{The baseline architectures are defined as \emph{Conv-AE}, \emph{Pooling-AE}, and \emph{Dense-AE}. The following operations are involved: C = Convolution, US~=~Upsampling, MP = Max Pooling, F = Flatten, R = Reshape. $d$ is the image dimension: $d=4$ for SWIR and $d=3$ for laser samples.}
    \label{fig:pad_scheme}
\end{figure*}
\subsection{PAD Scheme}
\label{sec:pad_scheme}

We study three different architectures of an AE, as illustrated in Fig.~\ref{fig:pad_scheme}, in order to find the best suited approach for fingerprint PAD. The four SWIR images are concatenated to a single 4-channel image such that one AE can work on all information simultaneously. Taking the first, middle, and last frame of the laser sequence, a second AE is trained on a 3-channel input image. 
In contrast to a LSTM~\cite{lstm}, the AE is not designed to learn temporal correlation, and since the changes within this sequence are subtle, we decided to take into account only the three furthermost frames in a similar way as the SWIR images are used. 
Due to the hardware changes of the capture device, computing the contrast of the laser sequence data does not work anymore as opposed to previous work~\cite{Keilbach-Fingerprint-LSCI-PAD-BIOSIG-2018, Kolberg-LSCIBenchmarkFingerPAD-BIOSIG-2019}. Hence, we discard the term LSCI and refer to laser sequences (or laser) in this work.

We denote the three architecture types as \textbf{Conv-AE}, \textbf{Pooling-AE}, and \textbf{Dense-AE} (top to bottom in Fig.~\ref{fig:pad_scheme}). The names refer to the type of layers which were successively added to the architecture. The \textbf{Conv-AE} is composed of convolutional layers with a stride value of two in order to reduce the dimension during the encoding phase. In the \textbf{Pooling-AE}, the stride value of the convolutional operations was changed to one, followed by a max pooling operation to reduce the dimension. The last modification \textbf{Dense-AE} added a Fully Connected Neural Network (Fully-Connected NN) between the encoding and decoding phase to reduce the dimension of the original image down to a 64-dimensional vector. 
All baseline architectures include a single convolutional / max pooling layer in the encoding phase.

The distinction between the Conv- and Pooling AE is grounded on the findings of Springenberg \etal~\cite{springenberg2014striving}, who claim that the max pooling operation can simply be replaced by a convolutional layer with an increased stride without significant loss in accuracy. On the other hand, Goodfellow \etal~\cite{goodfellow2016deep} state that the max pooling operation leads to an invariance of translations in smaller regions. Finally, the Dense-AE is inspired by Ke \etal~\cite{ke2017anomaly}, who emphasise the ability of the Fully-Connected NN to combine local features and to find interdependent patterns within the feature maps. 
Across all architectures the \emph{relu} activation function is used in all layers except for the very last convolutional layer, where the \emph{sigmoid} function proved to be the better choice. The convolutional layer includes twelve filters and MSE (Eq.~\eqref{eq:mse}) is used to measure the reconstruction error. 

In a second step, we evaluate the influence of the reconstruction error. In particular, we take the best-performing architecture and compare the MSE approach to the wMSE approach by analysing different constant values $C$ for the threshold computation. Hence, for each adaptation a new model is trained, since the loss function changes the learned weights during training.

Finally, we are interested in the best fusion of both AE types, based on SWIR and laser data, since previous approaches~\cite{Hussein-LSCI-SWIR-CNN-FingerPAD-WIFS-2018, Gomez-Barrero-FusionBATL-PAD-UBIO-2018, GomezBarrero-PAD-SWIR-LSCI-ICB-2019} show a significant improvement in PAD performance. For this reason, we compute different weighted fusions and compare the results in order to find the one best suited for our fingerprint PAD approach.

	\section{Experimental Evaluation}
	\label{sec:eval}
	Starting with a description of the utilised dataset and protocol, this Section provides the details of the experimental evaluation. Subsequently, the results of our PAD method are presented and finally benchmarked with additional classifiers.
\subsection{Database and Experimental Protocol}

\begin{table}[t]
\small
	\centering
	\setlength{\tabcolsep}{3pt}
	\caption{Summary of PAIs in the database with their corresponding group. The number of total samples and the number of variations is given. Variations include e.g.\ different colours and conductive augmentations.}
	\label{tab:pais}
	\begin{tabular}{llcc}
		\toprule
		\textbf{PAI Group} & \textbf{PAI} & \textbf{\#variations} & \textbf{\#samples} \\ 
		\midrule
		\multirow{8}{*}{Fakefinger} & 3D printed & 2 & 72 \\ 
		& dental material & 1 & 33 \\ 
		& dragon skin & 3 & 477 \\ 
		& ecoflex & 4 & 291 \\ 
		& latex & 2 & 147 \\ 
		& playdoh & 4 & 116 \\ 
		& silly putty & 3 & 55 \\ 
		& wax & 1 & 74 \\ 
		\midrule
		\multirow{8}{*}{Overlay opaque} & bandage plaster & 1 & 14 \\ 
		& dental material & 1 & 51 \\ 
		& dragon skin & 1 & 17 \\ 
		& ecoflex & 2 & 1035 \\ 
		& gelatin & 1 & 194 \\ 
		& printout paper & 1 & 49 \\
		& silicone & 4 & 752 \\ 
		& urethane & 1 & 72 \\ 
		\midrule
		\multirow{7}{*}{Overlay transparent} & dragon skin & 1 & 106 \\ 
		& gelatin & 1 & 107 \\ 
		& glue & 2 & 27 \\ 
		& latex & 1 & 34 \\ 
		& printout foil & 1 & 64 \\
		& silicone & 1 & 157 \\ 
		& wax & 1 & 18 \\ 
		\midrule
		\multirow{4}{*}{Overlay semi} & dragon skin & 1 & 47 \\ 
		& ecoflex & 1 & 24 \\ 
		& glue & 2 & 146 \\ 
		& silicone & 1 & 160 \\ 
		\bottomrule
	\end{tabular} 
\end{table}

The data was collected in four acquisition sessions in two distinct locations within a timeframe of nine months. Subjects could participate multiple times and presented six to eight fingers per capture round including thumb, index, middle, and ring fingers.
Fingers were presented as they were, which resulted in samples with different levels of moisture, dirt, or ink. Further details about the capture process are given in \cite{Spinoulas-BATLDatasetFingerPAD-Arxiv-2020}.
The combined database contains a total of 24,050 samples comprising 19,711 bona fides and additional 4,339 PAs stemming from 45 different PAI species. These PAI species include full fake fingers and more challenging overlays as summarised in Table~\ref{tab:pais}. The printouts were also worn as overlays and conductive paint was applied to some PAIs.
Note that the project sponsor indicated to make the complete dataset available in the near future for reproducibility and comparison\footnote{\url{https://www.isi.edu/projects/batl/data}}.

\begin{table}[t]
	\centering
	\caption{Partition of training, validation, and test datasets.}
	\label{tab:partitioning}
	\begin{tabular}{lccc}
		\toprule
		&  Samples &  BF Samples &  PA Samples\\
		\midrule
		Training set & ~5,717 & ~5,717 & 0 \\
		Validation set & ~3,553 & ~3,553 & 0 \\
		Test set & 14,780 & 10,441 & 4,339 \\
		\midrule
		Total & 24,050 & 19,711 & 4,339 \\
		\bottomrule
	\end{tabular}
\end{table}
The combined database is split into non-overlapping training, validation, and test sets, where subjects who participated multiple times are included in only one of the sets. This ensures a fair evaluation on unseen samples at the test stage. Randomly assigning 30\% of the subjects to the training and additional 20\% to the validation set results in the partitioning shown in Table~\ref{tab:partitioning}.

Our implementation is done with Keras~\cite{chollet2015keras}, which is a python based deep learning library that facilitates the definition, training and evaluation of various deep learning model types. For training the parameters, we used the pre-implemented \emph{RMSprop}~\cite{rmsprop} as an adaptive optimiser.

The PAD performance is shown in detection error trade-off (DET) curves between the BPCER and the APCER. For further comparison the partial area under curve (pAUC) of up to 20\% error rate is computed for each curve. It should be noted that the PAD threshold can be adjusted depending on the use case: A low BPCER represents a very convenient system, while a low APCER is more important for high security applications. Furthermore, the detection equal error rate (D-EER) is the point where APCER = BPCER.

\begin{figure}[t]
    \centering
    \includegraphics[width=\linewidth]{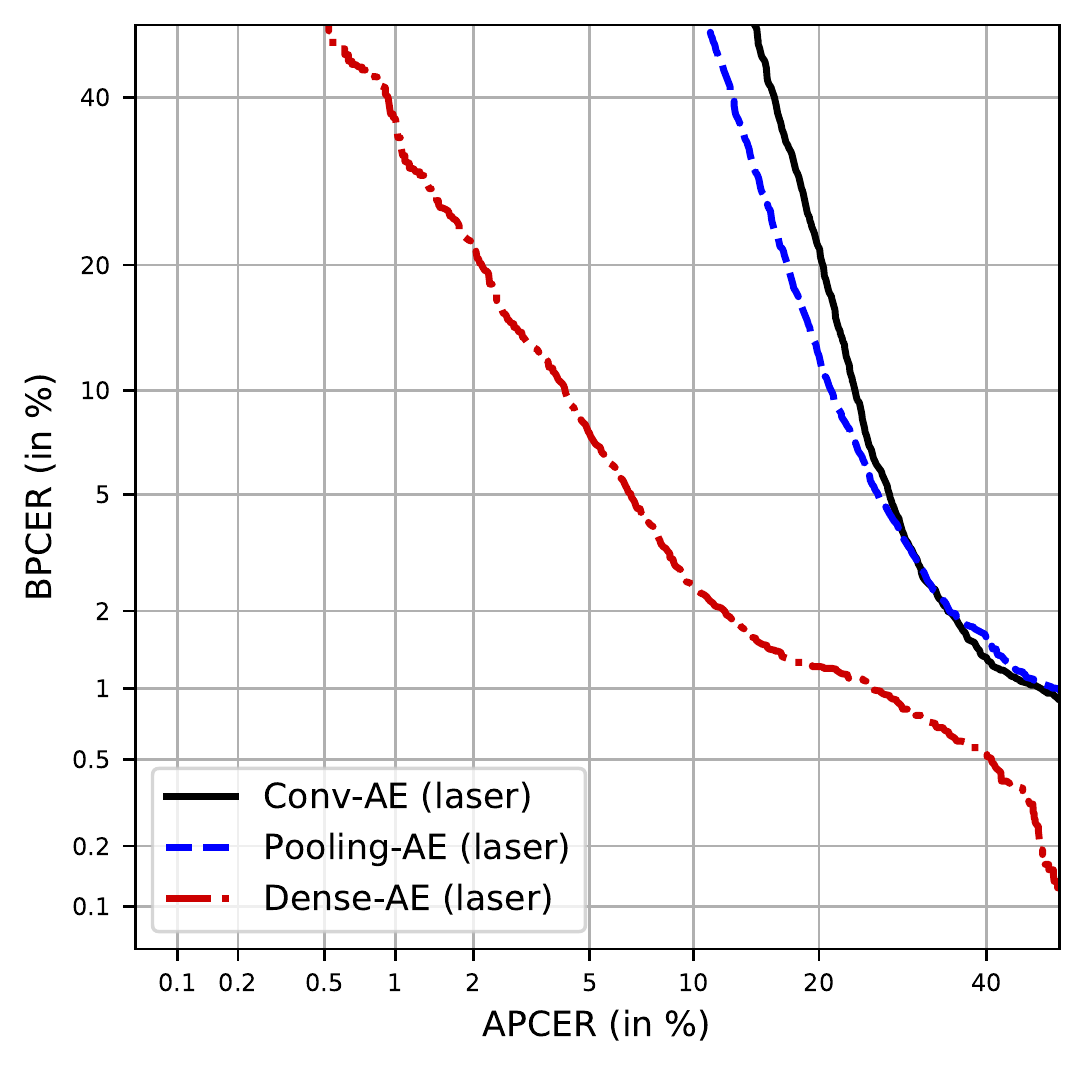}
    \\
    \includegraphics[width=\linewidth]{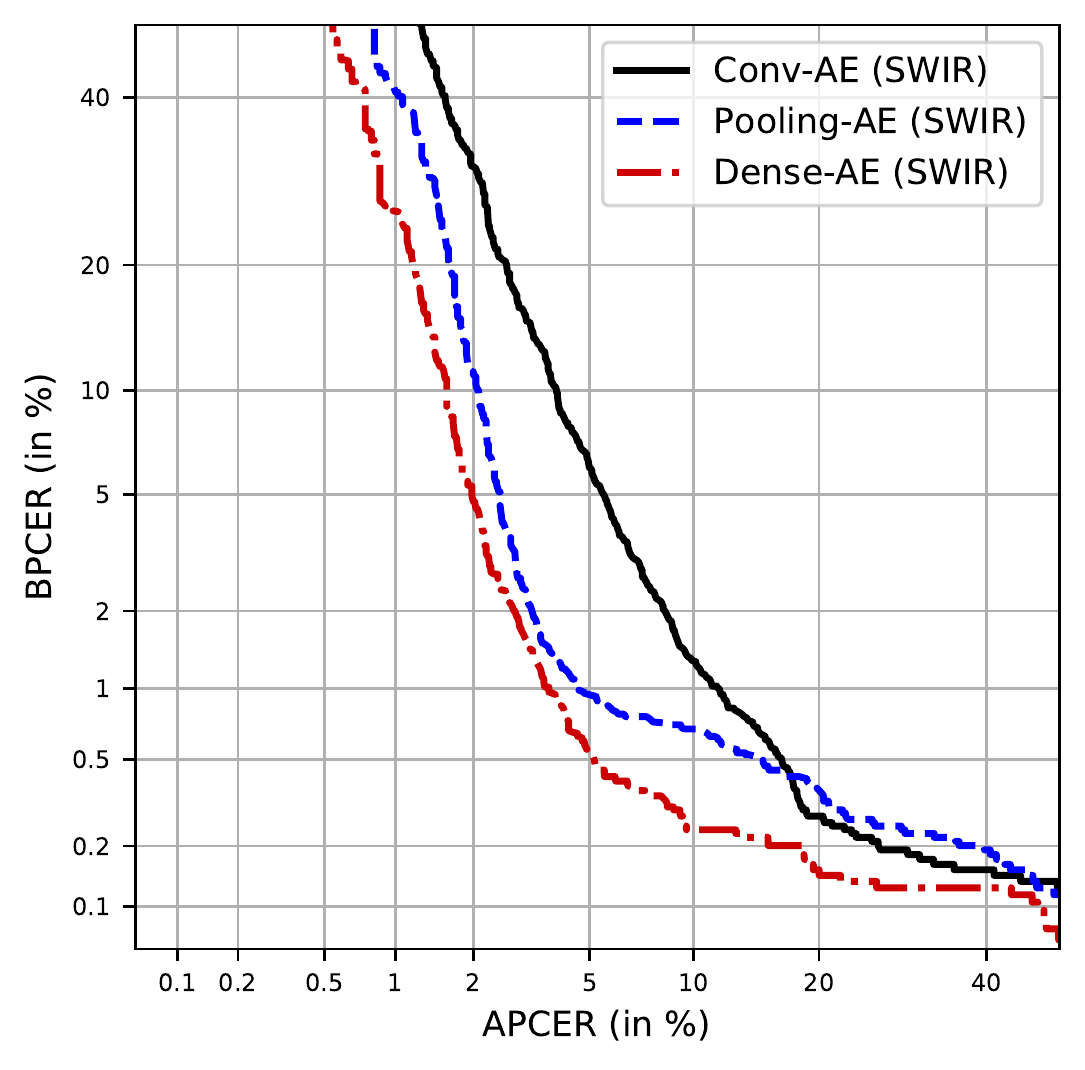}
    \caption{DET curves for the three evaluated AE architectures on laser (top) and SWIR (bottom) data.}
    \label{fig:det_structure}
\end{figure}

\begin{figure}
    \centering
    \includegraphics[width=\linewidth]{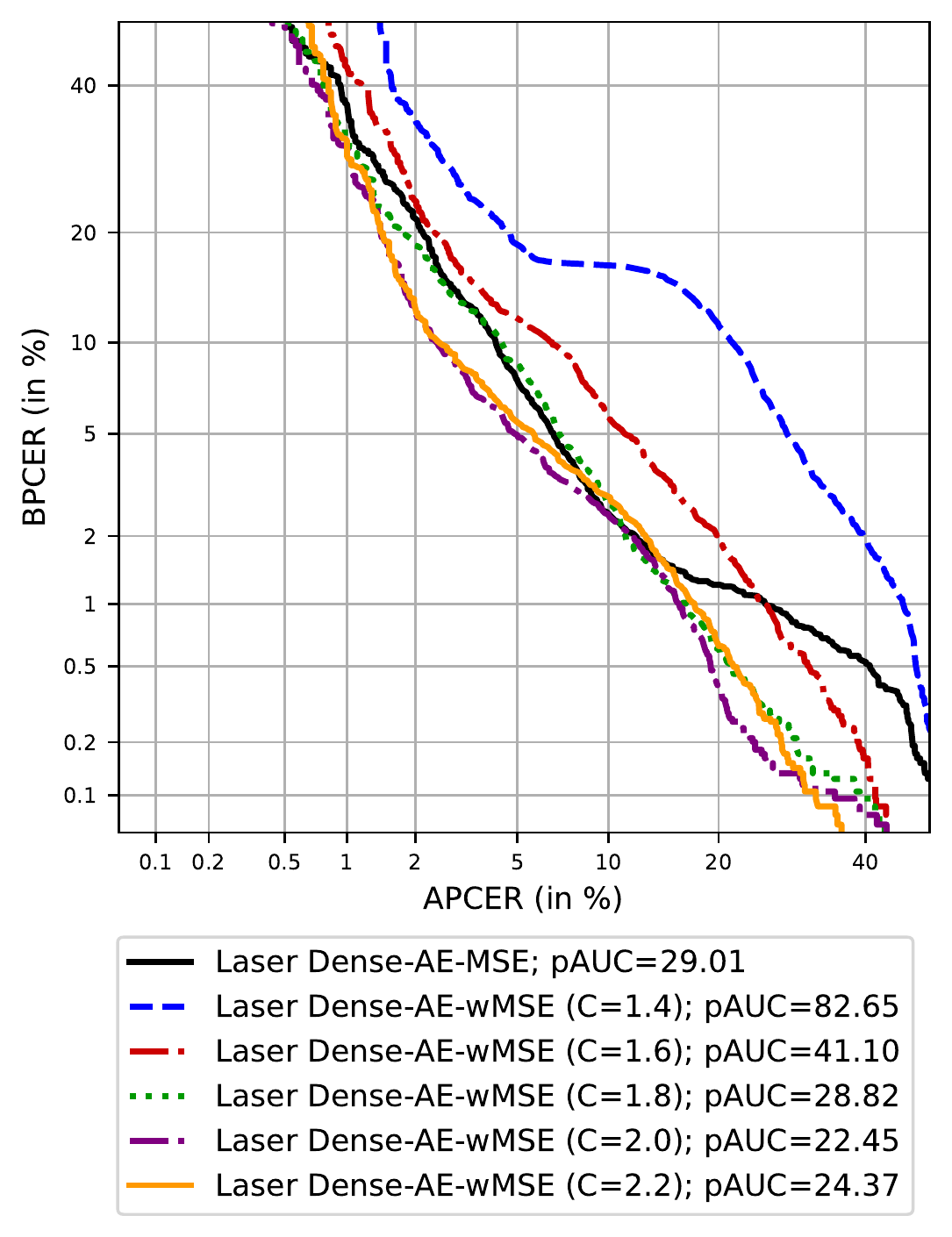}
\\
    \includegraphics[width=\linewidth]{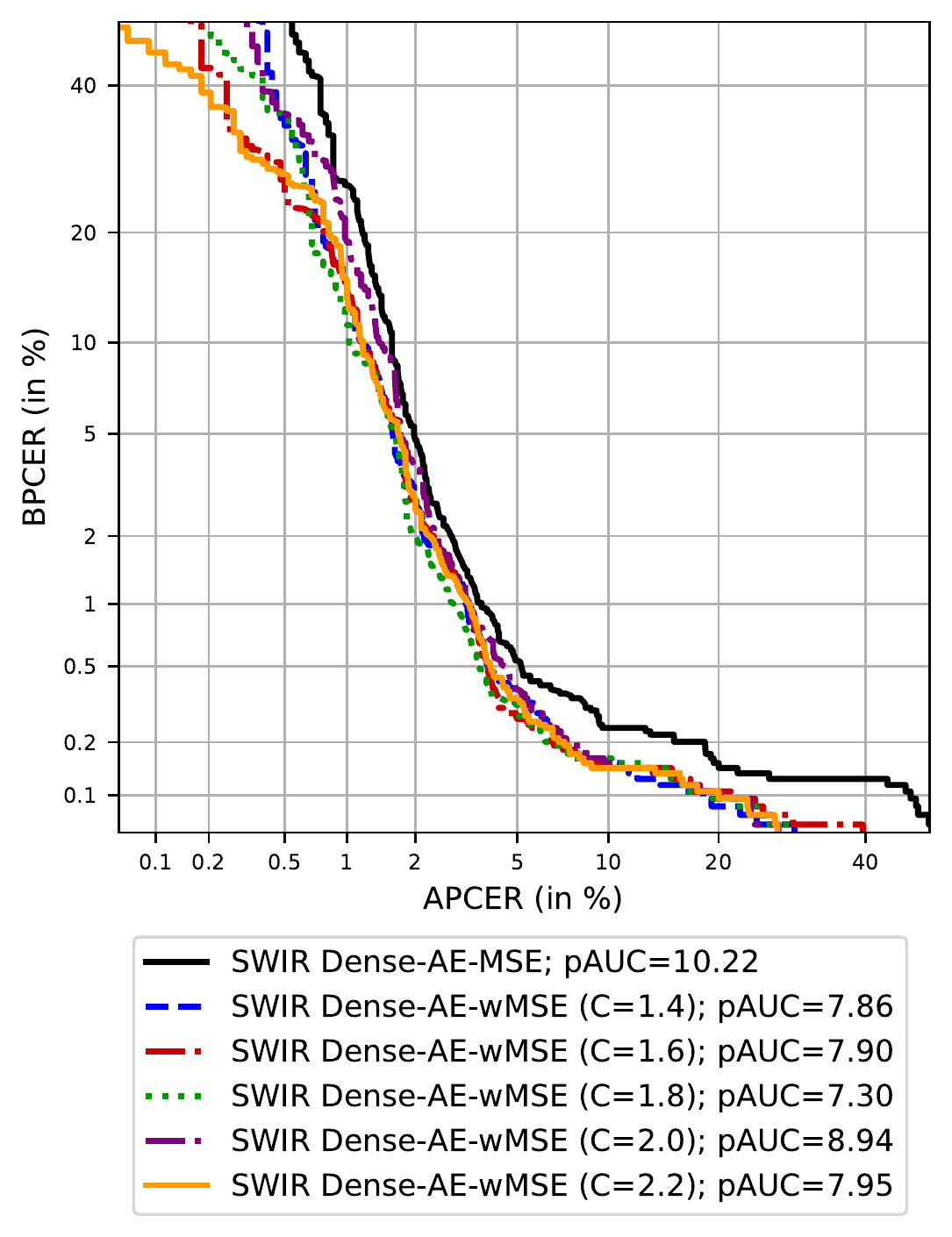}
    \caption{DET curves and pAUC (\%) of the Dense-AE comparing the MSE approach to the wMSE optimisation on laser (top) SWIR (bottom) data.}
    \label{fig:det_wmse}
\end{figure}

\subsection{PAD Method Evaluation}

The first part of our experiments compares the three baseline architectures: Conv-AE, Pooling-AE, and Dense-AE. The corresponding DET curves for both laser (top) and SWIR (bottom) input data are shown in Fig.~\ref{fig:det_structure}. In both cases, the Dense-AE (red) achieves the best performance at all thresholds. Therefore, it can be concluded that the Dense-AE is better capable of extracting relevant latent features of the given input data, that can be reconstructed to the original image.

\begin{figure}
    \centering
    \includegraphics[width=\linewidth]{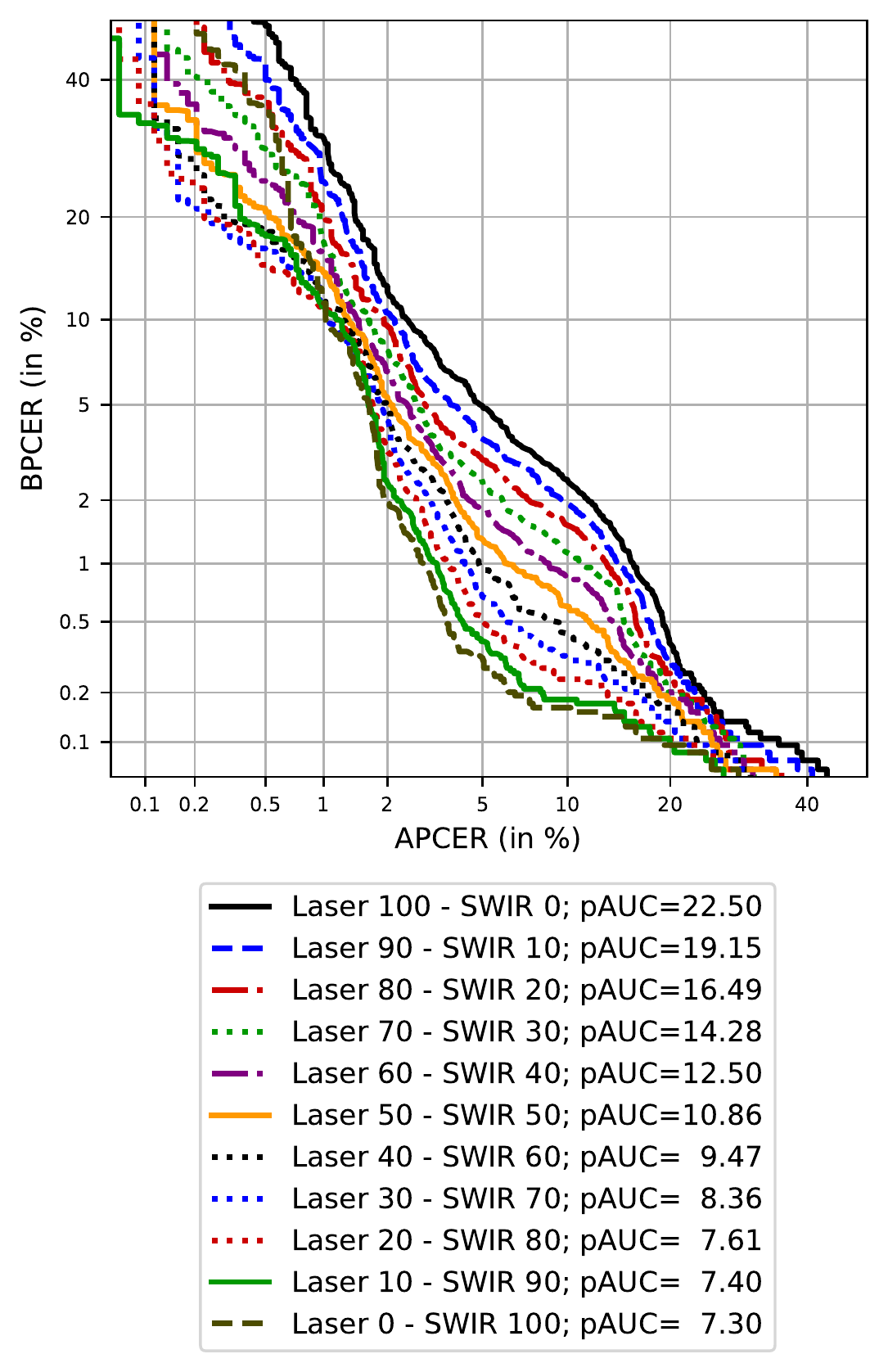}
    \caption{DET curves and pAUC (\%) from different weighted score-level fusions of the best-performing wMSE Dense-AEs.}
    \label{fig:det_fusion}
    \vspace*{-0.2cm}
\end{figure}

In the next step, the MSE (Eq.~\ref{eq:mse}) has been replaced by our proposed wMSE (Eq.~\ref{eq:proposed_wmse}). Since the wMSE involves another hyperparameter $C$, Fig.~\ref{fig:det_wmse} depict the DET curves for different parameter choices for laser and SWIR data, respectively. Also, the best performing baseline model has been added (Dense-AE with MSE) in order to directly compare it with the new settings. Looking at the graphs and the pAUC values, the performance increases for growing values of $C$. This indicates that by choosing $C$ too low, the excluded image areas are too large, which in turn leads to a loss of information. This phenomenon can be observed up to values of $C$=$2.2$ (laser) and $C$=$2.0$ (SWIR), where the performance decreases again. Choosing $C$ values that are too high leads to thresholds, that non of the pixel-wise REs exceed. Therefore, too few areas are excluded from the training process. 
Hence, in our experiments, values of $C$=$2.0$~(laser) and $C$=$1.8$ (SWIR) proved to be good choices.   

To evaluate whether the laser and SWIR AE models complement each other, we applied a weighted score fusion and the resulting DETs are depicted in Fig.~\ref{fig:det_fusion}. The given pAUCs show that the performance constantly decreases for higher weights on the laser scores. Thus, the optimal setting is to only use the SWIR scores as any inclusion of the laser scores has a negative effect on the classification results. On the other hand, for a possible high security application (e.g., APCER = 0.1\%) the fusion benefits from the laser-based PAD. However, the BPCER values are above our 20\% pAUC mark and thus not considered in computing the pAUC.

When analysing the occurring APCEs for a convenient BPCER=0.2\%, we found that all falsely classified PA samples of the SWIR AE are also misclassified by the laser AE. This includes mostly transparent overlays of clear dragon skin and two part silicone or full finger PAIs in yellow and orange playdoh. Also previous works~\cite{Gomez-Barrero-SWIR-SS-PAD-NISK-2018, GomezBarrero-MSSWIRCNN-CRC-2020} on SWIR PAD had troubles with orange playdoh since its reflections are nearly identical to skin within the SWIR spectrum. The other APCEs are still close enough to bona fide representations that the reconstruction errors could not be distinguished.
In addition to the already mentioned APCEs, the laser AE further fails to detect full finger PAIs of dragon skin, ecoflex, and monster latex as well as overlays out of gelatin, school glue, ecoflex, gelatin, and monster latex. Since the laser samples are all captured in the same wavelength, PAIs are more likely to resemble bona fide samples.

\begin{figure}[t]
    \centering
    \input{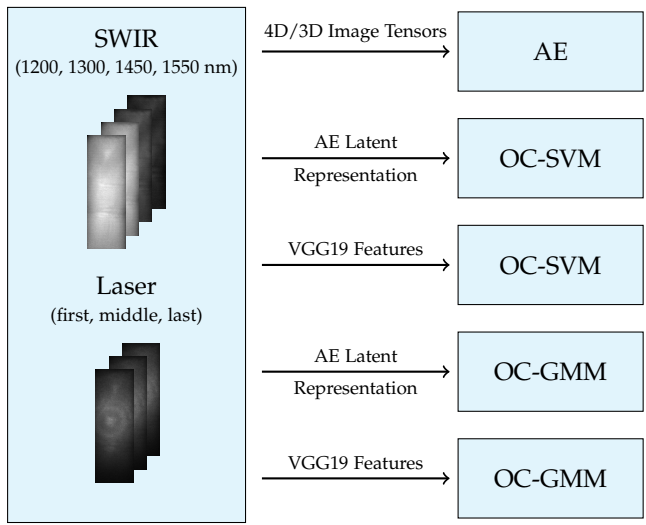}
    \caption{Overview of the additionally benchmarked one-class classifiers and their corresponding input features.}
    \label{fig:benchmark_structure}
\end{figure}

\subsection{Benchmark with other One-class Classifiers}

\begin{figure}
    \centering
    \includegraphics[width=\linewidth]{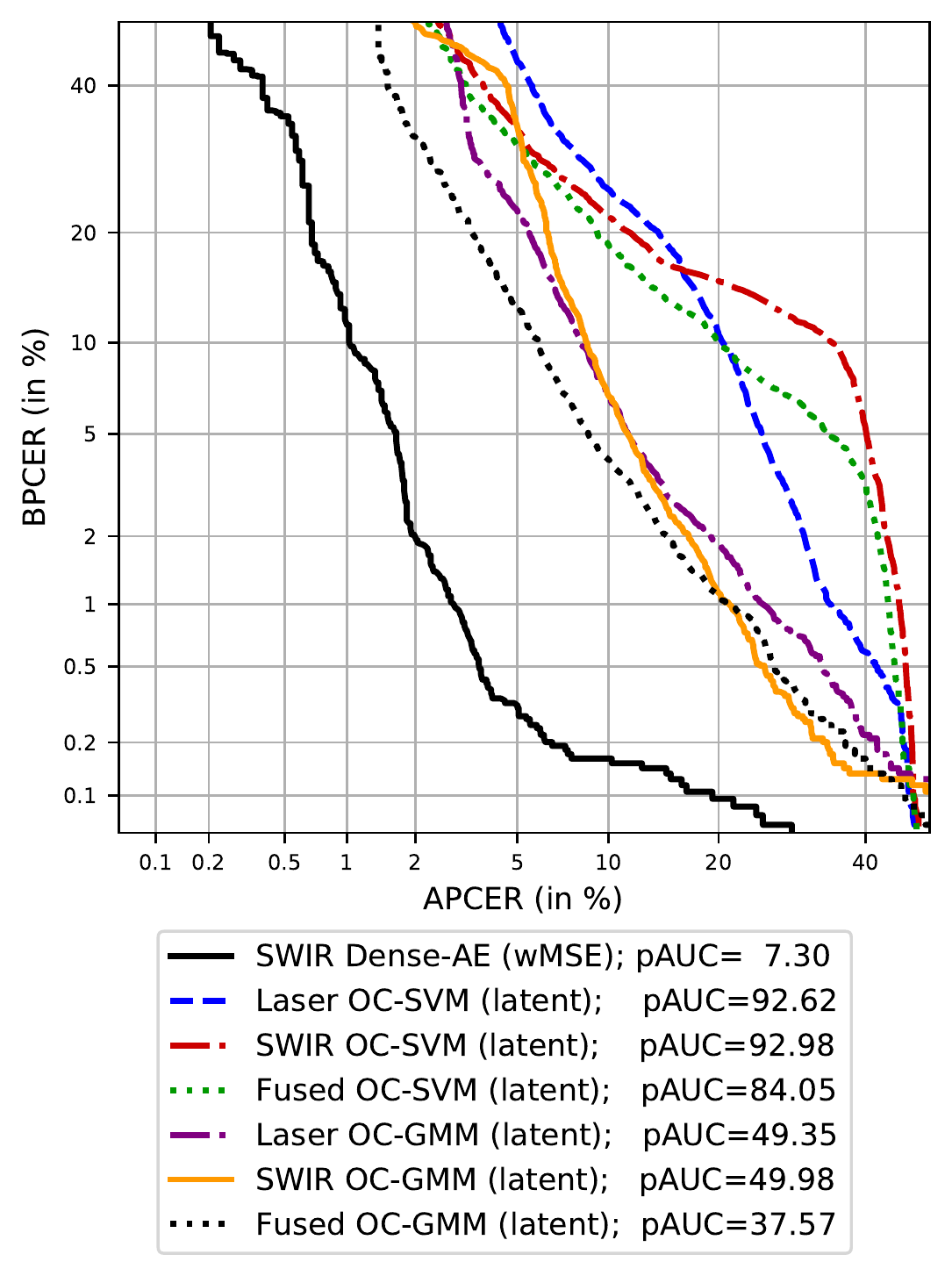}
    \caption{DET curves and pAUC (\%) from the benchmark of SWIR wMSE Dense-AE to other classifiers trained on latent AE representations.}
    \label{fig:det_benchmark_latent}
\end{figure}

\begin{figure}
    \centering
    \includegraphics[width=\linewidth]{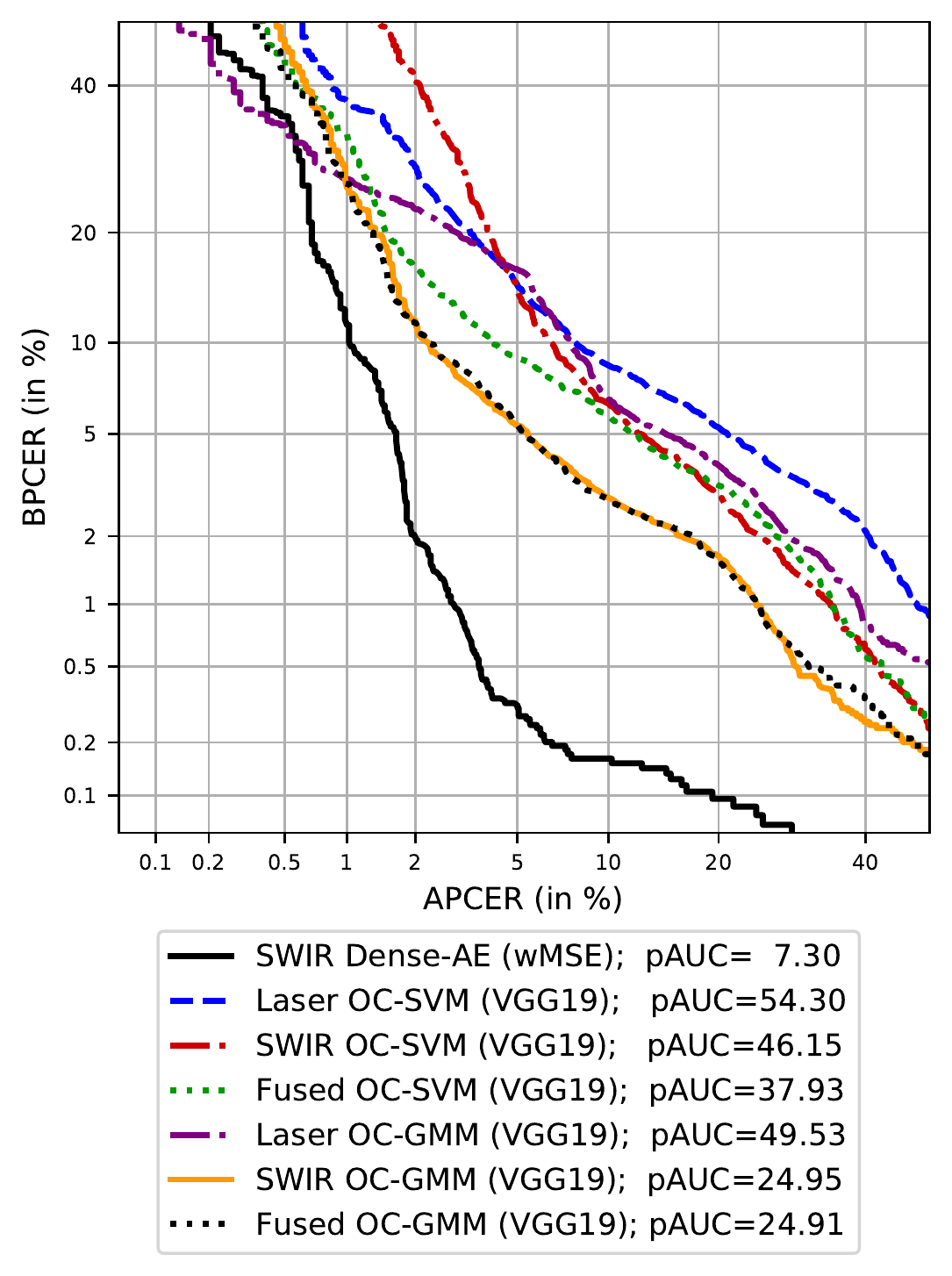}
    \caption{DET curves and pAUC (\%) from the benchmark of SWIR wMSE Dense-AE to other classifiers trained on features extracted by VGG19.}
    \label{fig:det_benchmark_vgg19}
\end{figure}

Summarising the results so far, the best performance could be obtained with the Dense-AE trained on the SWIR dataset using the proposed wMSE. To put these numbers into context, we benchmark our proposed AE with further one-class classifiers. In this context, we train and test an \mbox{OC-SVM}~\cite{ocsvm} and an OC-GMM~\cite{gmm} on two different feature representations of the input images. One is the latent feature representation as a result of the encoding phase from our Dense-AE and the other method utilises the VGG19~\cite{vgg} CNN pre-trained on \cite{Kolberg-LSTM-FingerPAD-IJCB-2020} to only extract features from the given input. This results in a total of four combinations of classifiers and features for each SWIR and laser data as depicted in Fig.~\ref{fig:benchmark_structure}. Finally, the laser and SWIR approaches are also fused to enhance their detection accuracy.
Fig.~\ref{fig:det_benchmark_latent} and Fig.~\ref{fig:det_benchmark_vgg19} visualise how the AE benchmarks against other one-class classifiers. The first graph contains the performance of OC-SVMs and OC-GMMs trained on the latent representations of the AE. The second graph shows the DET curves of both classifiers trained on features extracted with a pre-trained CNN (see Section~\ref{sec:pad}). The AE performs significantly better than both other approaches since its curves are well below the other methods. Interestingly, the fused OC-GMM performs second-best with a pAUC of 37.57\% (latent) and 24.91\% (VGG19). Contrary to the AE, the performances of the OC-SVMs and OC-GMMs can be improved by fusing the laser and SWIR scores. Thus, in contrast to the AE, there is a complementary effect measurable. 

In order to connect our results with the state-of-the-art, Table~5 shows the APCER values of this final benchmark for a very convenient BPCER of 0.2\%. Furthermore, the D-EER is depicted as the point where APCER = BPCER. While above pAUC evaluated the general performance, those two operation points (APCER at BPCER=0.2\%, D-EER) suit better to report specific results. The values show that both Dense-AEs outperform the other classifiers and prove the superiority of our SWIR Dense-AE. While our method achieves an APCER of 6.59\% at a BPCER of 0.2\% and a D-EER of 2.00\%, the best performance of all other classifiers is an APCER of 34.17\% at a BPCER of 0.2\% for the SWIR OC-GMM based on the latent representation, and a D-EER of 5.16\% for the fused OC-GMM based on the VGG19 representation.

\begin{table}[t]
    \centering
    \setlength{\tabcolsep}{8pt}
    \caption{Overview of APCERs for a fixed BPCER of 0.2\% and D-EERs.}
    \label{tab:results}
    \begin{tabular}{lrr}
    \toprule
    Algorithm & APCER & D-EER \\
    \midrule
    Laser Dense-AE (wMSE) & 24.33\% & 4.96\% \\
    SWIR Dense-AE (wMSE) & \textbf{6.59\%} & \textbf{2.00\%} \\
    \midrule
    Laser OC-GMM (latent) & 41.25\% & 8.80\% \\
    SWIR OC-GMM (latent & 34.17\% & 8.98\% \\
    Fused OC-GMM (latent) & 36.88\% & 7.09\% \\
    Laser OC-SVM (latent) & 45.95\% & 16.21\% \\
    SWIR OC-SVM (latent) & 47.27\% & 16.18\% \\
    Fused OC-SVM (latent) & 45.86\% & 13.93\% \\
    \midrule
    Laser OC-GMM (VGG19) & 63.98\% & 8.64\% \\
    SWIR OC-GMM (VGG19) & 47.43\% & 5.21\% \\
    Fused OC-GMM (VGG19) & 47.41\% & 5.16\% \\
    Laser OC-SVM (VGG19) & 65.12\% & 9.07\% \\
    SWIR OC-SVM (VGG19) & 52.82\% & 7.96\% \\
    Fused OC-SVM (VGG19) & 55.55\% & 7.17\% \\
    \bottomrule
    \end{tabular}
\end{table}
	
	\section{Conclusion}
	\label{sec:conclusion}
	In this paper, we have proposed a one-class PAD method based on convolutional autoencoders. Specifically, we compared three different AE architectures (Conv-AE, Pooling-AE, and Dense-AE). Based on our experiments, we can conclude that the Dense-AE performs significantly better than the other model architectures on both laser and SWIR input images.

Additionally, we proposed the wMSE as an extension of the MSE with the idea of ignoring disturbing image areas (e.g.\ illumination interference) during both training and testing. With the MSE replaced by the wMSE, the pAUC values could further be improved from 29.01\% to 22.45\% (laser) and from 10.22\% to 7.30\% (SWIR). The weighted fusion of the laser and SWIR scores did not improve the results. Therefore, in contrast to related work applying two-class approaches, the two AEs do not complement each other. 

Finally, two additional well-established one-class classifiers (OC-SVMs and OC-GMMs) have been trained on two different feature inputs. The first set of OC-SVMs and OC-GMMs were trained on the latent representations of the best performing AE. The second features have been extracted with a VGG19~\cite{vgg} CNN pre-trained on \cite{Kolberg-LSTM-FingerPAD-IJCB-2020}. None of the alternative one-class classifiers achieved a comparable performance to our proposed Dense-AE, which proves the soundness of the approach. Nevertheless, both alternative methods benefit from information fusion of laser and SWIR data.

Future work will focus on further optimising the wMSE. In our work, every pixel-wise RE gets an individual weight (zero or one) depending on whether it exceeds the chosen threshold $C$ or not. This binary criterion could be loosened to allow the weights to have values between zero and one. 
Additionally, the concept of the Dense-AE can be applied to further PAD tasks as face and iris PAD, or software-based fingerprint PAD on the LivDet datasets.

	
	%

	
	
	\ifCLASSOPTIONcompsoc
	\section*{Acknowledgments}
	\else
	\section*{Acknowledgment}
	\fi
	This research is based upon work supported in part by the Office of the Director of National Intelligence (ODNI), Intelligence Advanced Research Projects Activity (IARPA) under contract number 2017-17020200005. The views and conclusions contained herein are those of the authors and should not be interpreted as necessarily representing the official policies, either expressed or implied, of ODNI, IARPA, or the U.S.\ Government. The U.S.\ Government is authorized to reproduce and distribute reprints for governmental purposes notwithstanding any copyright annotation therein.\\
	This research work has been funded by the German Federal Ministry of Education and Research and the Hessen State Ministry for Higher Education, Research and the Arts within their joint support of the National Research Center for Applied Cybersecurity ATHENE.\\
	Furthermore, we would like to thank our colleagues from Information Science Institute at University of Southern California for the data collection effort.

	\ifCLASSOPTIONcaptionsoff
	\newpage
	\fi

	
	
	\bibliographystyle{IEEEtran}
	\bibliography{IEEEabrv,ae_bib}
	%

	%
	
	\begin{IEEEbiography}[{\includegraphics[width=1in,height=1.25in,clip,keepaspectratio]{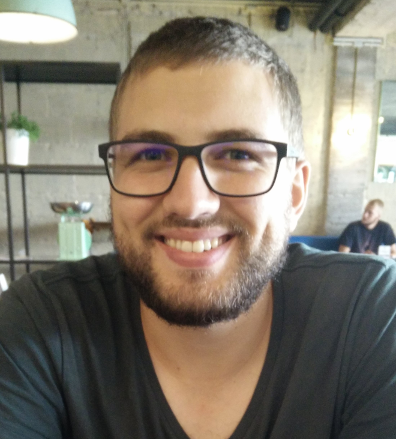}}\vspace*{10pt}]{Jascha Kolberg}
		received his B.Sc.\ and M.Sc.\ degrees in IT-Security / Information Technology from Ruhr-University Bochum in 2014 and 2017, respectively. Since 2017 he is a Ph.D.\ Student Member of da/sec at the National Research Center for Applied Cybersecurity (ATHENE). His current research focuses on biometric template protection and presentation attack detection for fingerprint recognition sensors.
	\end{IEEEbiography}
	
	\begin{IEEEbiography}[{\includegraphics[width=1in,height=1.25in,clip,keepaspectratio]{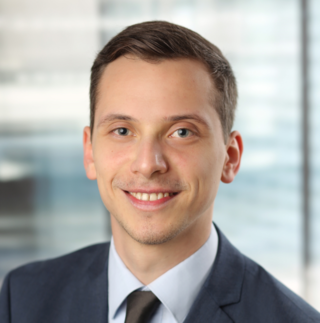}}]{Marcel Grimmer}
		received his B.Sc.\ and M.Sc.\ degrees in Applied Mathematics and Data Science from University of Applied Sciences Darmstadt in 2018 and 2020. Since 2020 he is working as a Ph.D.\ Student of the Norwegian Biometrics Laboratory at the Norwegian University of Science and Technology (NTNU). During his studies, his research focused on presentation attack detection for fingerprints. Since July, 2020 his main research is dedicated to image quality assessment of facial images.
	\end{IEEEbiography}
	
	
	\begin{IEEEbiography}[{\includegraphics[width=1in,height=1.25in,clip,keepaspectratio]{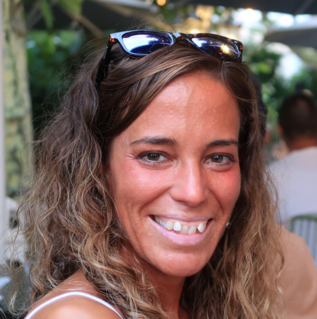}}]{Marta Gomez-Barrero}
		is a Professor for IT-Security and technical data privacy at the Hochschule Ansbach, in Germany. Between 2016 and 2020, she was a postdoctoral researcher at the National Research Center for Applied Cybersecurity (ATHENE) - Hochschule Darmstadt, Germany. Before that, she received her M.Sc.\ degrees in Computer Science and Mathematics (2011), and her Ph.D.\ degree in Electrical Engineering (2016), all from Universidad Autonoma de Madrid, Spain. Her current research focuses on security and privacy evaluations of biometric systems, presentation attack detection methodologies, and biometric template protection schemes. She has co-authored more than 70 publications, chaired special sessions and competitions at international conferences, she is associate editor for the EURASIP Journal on Information Security, and represents the German Institute for Standardization (DIN) in ISO/IEC SC37 JTC1 SC37 on biometrics. She has also received a number of distinctions, including: EAB European Biometric Industry Award 2015, Best Ph.D.\ Thesis Award by Universidad Autonoma de Madrid 2015/16, Siew-Sngiem Best Paper Award at ICB 2015, Archimedes Award for young researches from Spanish MECD, and Best Poster Award at ICB 2013.
	\end{IEEEbiography}
	\begin{IEEEbiography}[{\includegraphics[width=1in,height=1.25in,clip,keepaspectratio]{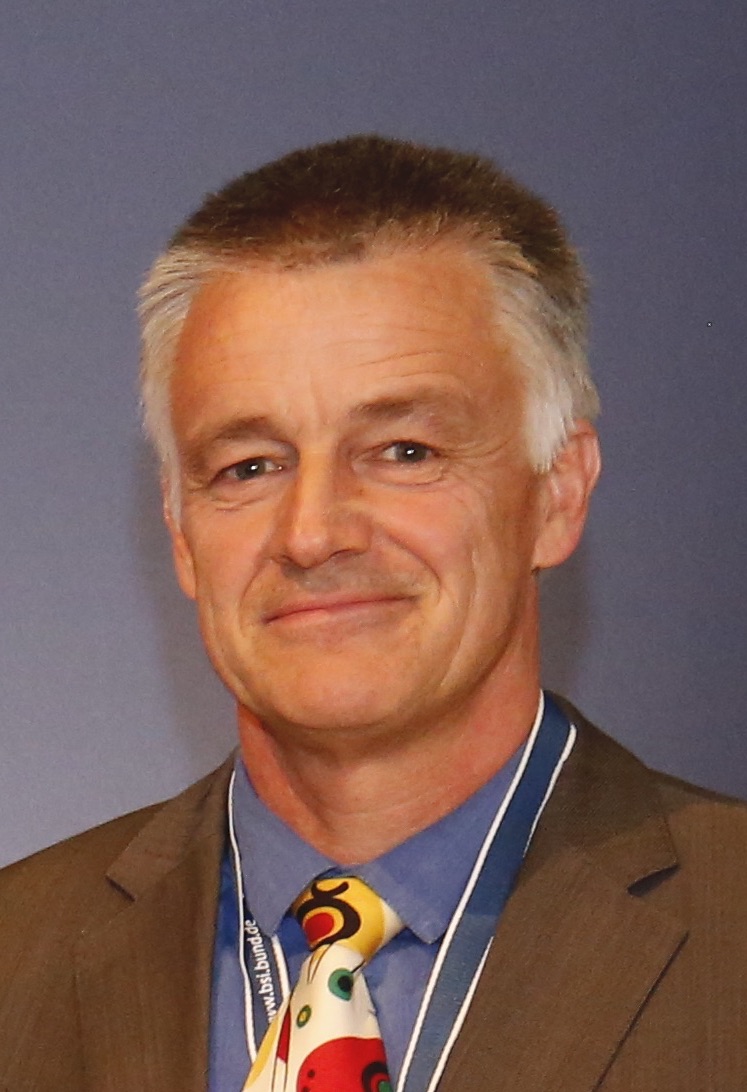}}]{Christoph Busch}
		received the Diploma degree from the Technical University of Darmstadt (TUD), Darmstadt, Germany, and the Ph.D.\ degree in computer graphics from TUD, in 1997. He joined the Fraunhofer Institute for Computer Graphics, Darmstadt, in 1997. He is a member of the Norwegian Biometrics Laboratory with the Norwegian University of Science and Technology, Norway, and holds a joint appointment with the Faculty of Computer Science, Hochschule Darmstadt. Furthermore, he lectures a course on biometric systems with DTU in Copenhagen since 2007. His research includes pattern recognition, multimodal and mobile biometrics, and privacy enhancing technologies for biometric systems. He is Cofounder of the European Association for Biometrics and convener of WG3 in ISO/IEC JTC1 SC37 on Biometrics. He coauthored over 500 technical papers, and has been a speaker at international conferences.
	\end{IEEEbiography}
	
	
	

\end{document}